\documentclass[sigconf]{acmart}
\AtBeginDocument{%
  }

\setcopyright{acmlicensed}
\copyrightyear{2018}
\acmYear{2018}
\acmDOI{XXXXXXX.XXXXXXX}
\acmConference[Conference acronym 'XX]{Make sure to enter the correct
  conference title from your rights confirmation email}{June 03--05,
  2018}{Woodstock, NY}
\acmISBN{978-1-4503-XXXX-X/2018/06}

\usepackage{algorithmic}
\usepackage{algorithm}
\usepackage{booktabs,multirow}
\usepackage[skip=5pt]{caption} 
\usepackage{graphicx} 

\begin{document}

\title{TIP: Resisting Gradient Inversion via Targeted Interpretable Perturbation in Federated Learning}

\author{Jianhua Wang}
\email{wangjianhua02@tyut.edu.cn}
\orcid{0000-0002-2773-3429}
\affiliation{%
  \institution{Shanxi Key Laboratory of Industrial Internet Security, College of Computer Science and Technology}
  \institution{Taiyuan University of Technology}
  \city{Taiyuan}
  \state{Shanxi}
  \country{China}
}

\author{Yilin Su}
\email{2024520738@link.tyut.edu.cn}
\orcid{0000-0000-0000-0000}
\affiliation{%
  \institution{Shanxi Key Laboratory of Industrial Internet Security, College of Computer Science and Technology}
  \institution{Taiyuan University of Technology}
  \city{Taiyuan}
  \state{Shanxi}
  \country{China}
}

\renewcommand{\shortauthors}{Wang et al.}

\begin{abstract}
Federated Learning (FL) facilitates collaborative model training while preserving data locality; however, the exchange of gradients renders the system vulnerable to Gradient Inversion Attacks (GIAs), allowing adversaries to reconstruct private training data with high fidelity. Existing defenses, such as Differential Privacy (DP), typically employ indiscriminate noise injection across all parameters, which severely degrades model utility and convergence stability. To address those limitation, we proposes \textbf{\textit{T}}argeted \textbf{\textit{I}}nterpretable \textbf{\textit{P}}erturbation (TIP), a novel defense framework that integrates model interpretability with frequency domain analysis. Unlike conventional methods that treat parameters uniformly, TIP introduces a dual-targeting strategy. First, leveraging Gradient-weighted Class Activation Mapping (Grad-CAM) to quantify channel sensitivity, we dynamically identify critical convolution channels that encode primary semantic features. Second, we transform these selected kernels into the frequency domain via the Discrete Fourier Transform and selectively inject calibrated perturbations into the high-frequency spectrum. By selectively perturbing high-frequency components, TIP effectively destroys the fine-grained details necessary for image reconstruction while preserving the low-frequency information crucial for model accuracy. Extensive experiments on benchmark datasets demonstrate that TIP renders reconstructed images visually unrecognizable against state-of-the-art GIAs, while maintaining global model accuracy comparable to non-private baselines, significantly outperforming existing DP-based defenses in the privacy-utility trade-off and interpretability. Code is available in https://github.com/2766733506/asldkfjssdf\_arxiv
\end{abstract}



\maketitle

\section{Introduction}
The rapid proliferation of Internet of Things (IoT) devices and edge computing has generated massive amounts of decentralized data. While this data holds immense value for training Deep Neural Networks (DNNs), traditional centralized training—which requires aggregating data on a single server—faces increasing scrutiny due to data privacy concerns and strict legal regulations such as the General Data Protection Regulation and the California Consumer Privacy Act. To address these challenges, Federated Learning (FL) \cite{mcmahan2017communication} has emerged as a promising paradigm. By enabling clients to collaboratively train a global model while keeping their raw data local, FL decouples data access from model training. The fundamental premise of FL is that exchanging model updates, e.g., gradients or weight parameters, is inherently safer than exchanging raw data. Consequently, FL has been widely deployed in privacy-sensitive applications, ranging from predictive text on mobile keyboards to medical image analysis and financial fraud detection.

\begin{figure}[h]
    \centering 
    \includegraphics[width=0.5\textwidth]{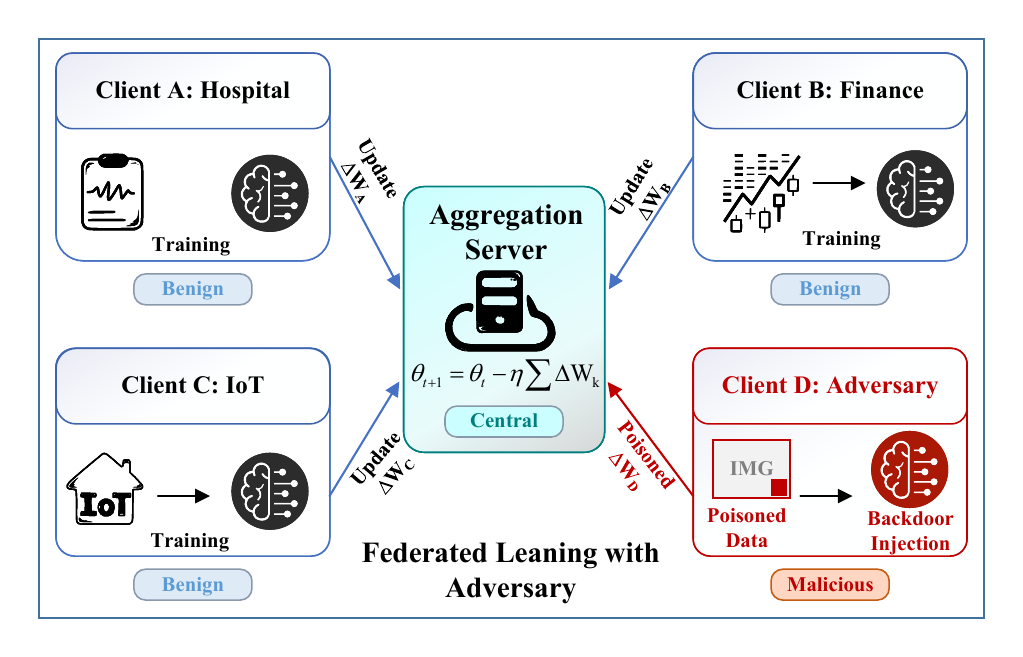}
    \caption{\centering The overall framework of the FL with Adversary}
    \label{fig:overall_FL}
\end{figure}

Nevertheless, the reliance on gradient sharing exposes FL to severe privacy risks that contradict its design goals. The assumption that gradients act as a secure firewall for raw data has been shattered by the emergence of deep leakage strategies. Since gradients dictate how a model should adjust to fit specific data points, they inadvertently leak sensitive patterns from the local datasets. This leakage provides a backdoor for Gradient Inversion Attack (GIA), where adversaries can mathematically recover the original training inputs from the shared updates. Such vulnerabilities reveal that without additional protection, the standard FL protocol fails to provide the strict privacy guarantees required by the very regulations it seeks to satisfy \cite{zhang2025secure}. Fig \ref{fig:overall_FL} demonstrates a FL framework with malicious clients.

Concretely, early attempts like Deep Leakage from Gradients (DLG) \cite{zhu2019deep} formulated reconstruction as an optimization problem, recovering inputs by minimizing the distance between dummy and real gradients. Building on this, advanced methods such as Inverting Gradients (IG) \cite{geiping2020inverting} have overcome initial limitations by leveraging cosine similarity metrics and total variation priors. These sophisticated techniques enable the recovery of private data with alarming detail—achieving pixel-perfect reconstruction even for complex images. Such findings provide empirical proof that standard FL protocols are insufficient to prevent white-box inference, necessitating the development of more rigorous privacy-preserving strategies.

While the threat of GIA is evident, establishing an effective defense mechanism is far from straightforward, as it requires navigating two fundamental \textbf{\textit{challenges}} inherent to distributed learning. 
\begin{itemize}
    \item \textbf{Privacy-Utility Trade-off.} Gradients serve as the essential carrier for model updates; therefore, any perturbation mechanism designed to obfuscate private data inevitably introduces noise that degrades the fidelity of the optimization direction \cite{Wei2020Fed, Kai2021Fed}. Achieving a strong privacy guarantee without causing significant deterioration in global model accuracy remains a central open problem.
    \item \textbf{Precise Leakage Localization}. In DNNs, the gradient vector often contains millions of parameters, where updates related to sensitive attributes are densely intermixed with benign, task-relevant updates. Existing literature indicates that without fine-grained interpretability, it is computationally prohibitive to distinguish which specific gradient elements contribute to privacy leakage versus those that are critical for model performance, often leading to suboptimal defense strategies that suppress valid learning signals \cite{zhu2019deep, Huang2021Adv}.
\end{itemize}
    
To mitigate these risks, current research has largely coalesced around Differential Privacy (DP)  \cite{abadi2016deep, geyer2017differentially} and Gradient Sparsification \cite{lin2017deep}, yet both approaches face critical \textbf{\textit{limitations}} when countering advanced reconstruction attacks.
\begin{itemize}
    \item \textbf{Severe Performance Degradation.} Differential Privacy mechanisms defend against inversion by injecting calibrated noise into the gradient updates. However, recent evaluations demonstrate that the noise scale required to theoretically thwart high-resolution reconstruction—especially against diffusion-based priors—often destroys the model’s convergence capabilities \cite{huang2021evaluating}. This forces a detrimental compromise where practitioners must accept either insufficient privacy protection or a model with drastically reduced utility \cite{balunovic2022bayesian}.
    \item \textbf{Lack of Semantic Awareness.} Gradient sparsification strategies attempt to reduce leakage by pruning parameters with small numerical magnitudes. However, these blind compression methods fail to distinguish between sensitive and non-sensitive attributes. Recent studies on adaptive attacks reveal that sensitive information is not strictly correlated with gradient magnitude, allowing adversaries to recover private data even from highly sparse updates \cite{fowl2021robbing, gong2023gradient, wang2024pa}. Without semantic guidance, these methods inadvertently discard valid learning signals while leaving critical leakage paths exposed.
\end{itemize}

To resolve those limitations, we propose a novel defense method against gradient inversion via \textbf{\textit{T}}argeted \textbf{\textit{I}}nterpretable \textbf{\textit{P}}erturbation in FL, namely \textbf{\textit{TIP}}. Our methodology is driven by two key \textbf{\textit{motivations}} that distinguish it from existing defenses:

\begin{itemize}
    \item \textbf{Interpretability-Guided Sensitivity.} Unlike conventional methods that treat all model parameters uniformly, we posit that parameters contribute heterogeneously to model inference. 
    By employing Gradient-weighted Class Activation Mapping (Grad-CAM) \cite{selvaraju2017grad} to calculate channel-wise importance weights, we can precisely isolate the critical convolution channels that encode primary semantic features. Empirical evidence suggests that applying perturbation to non-critical channels yields negligible privacy gains; therefore, an effective defense necessitates targeted obfuscation focused specifically on these high-contribution components to maximize the privacy-utility ratio.
    \item \textbf{Spectral Information Disentanglement.} From a signal processing perspective, there exists a distributional disparity between the information required for classification and the data exploited for reconstruction. Recent studies on the spectral bias of neural networks indicate that robust semantic features reside predominantly within low-frequency components. Conversely, the fine-grained details utilized by reconstruction attacks, including high-frequency textures and noise patterns, are embedded in the upper spectrum \cite{yin2019fourier}. This distinction allows us to decouple valid learning signals from leakage-prone artifacts and selectively suppress the latter in the frequency domain.
\end{itemize}

In this paper, based on these insights and motivations, TIP introduces a dual-targeting perturbation framework designed to reconcile the privacy-utility conflict. The methodology unfolds in two coordinated stages. First, we employ an interpretability-driven selection mechanism to identify convolution channels exhibiting high sensitivity to the target task. This step ensures that subsequent protective measures focus strictly on parameters containing dense semantic information. Second, the kernels of these selected channels undergo spectral transformation via the Two-Dimensional Discrete Fourier Transform (DFT). Diverging from conventional spatial noise injection, we implement a Targeted High-Frequency Injection strategy that introduces calibrated perturbations specifically into the high-frequency spectrum. This spectral manipulation effectively obfuscates the fine-grained gradient details exploited by adversaries for reconstruction while meticulously preserving the low-frequency structures indispensable for maintaining model utility and convergence stability.

The main contributions of this paper are summarized as follows:
\begin{itemize}
    \item \textbf{Interpretability-Driven Defense Framework.} We propose \textbf{\textit{TIP}}, the first FL defense paradigm that synergizes gradient-based interpretability with frequency domain analysis to mitigate GIAs. Unlike conventional methods that apply uniform noise, our framework dynamically identifies task-sensitive channels to shift the defense focus from blind perturbation to precision privacy preservation.
    \item \textbf{Frequency-Aware Perturbation Mechanism.} We devise a spectral injection algorithm that decomposes kernel weights via the DFT and selectively targets high-frequency components. This strategy theoretically disrupts the optimization landscape required for high-fidelity reconstruction while adhering to the spectral bias of neural networks to maintain the low-frequency structures essential for generalization.
    \item \textbf{State-of-the-Art Privacy-Utility Trade-off.} Extensive experiments on benchmark datasets validate that TIP renders reconstructed images visually unrecognizable effectively thwarting advanced inversion techniques. Our method significantly outperforms existing DP baselines by preserving global model accuracy comparable to non-private training scenarios establishing a new standard for balancing security and utility.
\end{itemize}

The rest of this paper is organized as follows. Section 2 outlines the necessary preliminaries. Section 3 details our proposed TIP. Section 4 presents the experimental setup. Finally, Section 5 concludes the paper.

\section{Preliminaries}
This section provides an overview of the fundamental concepts of FL, GIAs, and gradient-based interpretability techniques, which form the foundation for the work presented in this paper.

\subsection{Federated Learning}\label{sec:preli_fed}
A simple implementation of FL is as follows: First, the federated server distributes the global model $w_0$ to the clients. Then, 
the number of $j$ selected clients trains the model using their private data. Afterward, each client sends its locally trained model $w_j^t$ to the global server at the $t$-th communication round. Upon receiving the models uploaded by the clients, the server aggregates the weights from the clients to update and generate the new global model $w^t$. It can be expressed as equation \ref{eq:jvhe}.
\begin{equation}
\mathbf{w}^t \leftarrow \text{Aggregate}(\mathbf{w}_1^t, \dots, \mathbf{w}_j^t, \dots, \mathbf{w}_J^t)
\label{eq:jvhe}
\end{equation}

Subsequently, the federated server will distribute the new global model to the clients for the next round of training. The local update on each client is given by Equation \ref{eq:bendigengxin}.
\begin{equation}
\mathbf{w}_j^{t+1} \leftarrow \mathbf{w}^t - \eta \nabla F_j(\mathbf{w}_j^t)
\label{eq:bendigengxin}
\end{equation}

where $\eta$ is the learning rate and $\nabla F_j(\mathbf{w}_j^t)$ represents the gradient of the loss function for client $j$.

\subsection{Gradient Inversion Attacks}
GIA \cite{geiping2020inverting} pose a significant threat to client privacy in FL. In this paper, we assume that the federated server is honest-but-curious, strictly adhering to the FL protocol, while also analyzing the received information.

The general process of gradient attacks is as follows: first, a dummy input $x'$ and its corresponding label $y'$ are generated. Then, based on the model parameters $W$, the training process is executed to obtain the virtual gradient. It can be expressed as equation \ref{eq:xvnitidu}.
\begin{equation}
\nabla W' = \frac{\partial F \left( Y(x', W), y' \right)}{\partial W}
\label{eq:xvnitidu}
\end{equation}

Subsequently, iterative training is performed with the goal of minimizing the distance between the virtual gradient and the real gradient, as shown in equation \ref{eq:tidujvli}.
\begin{equation}
x'^{\ast}, y'^{\ast} = \arg\min_{x', y'} \quad Dist (\nabla W' , \nabla W )
\label{eq:tidujvli}
\end{equation}
The specific distance function includes equation \ref{eq:dist1} and equation \ref{eq:dist2}.
\begin{equation}
Dist (\nabla W' , \nabla W) = \left\| \frac{\partial F \left( Y(x', W), y' \right)}{\partial W} - \nabla W \right\|^2
\label{eq:dist1}
\end{equation}

\begin{equation}
Dist (\nabla W' , \nabla W) = 1 - \frac{\langle \nabla W, \nabla W' \rangle}{\|\nabla W\| \|\nabla W’\|}
\label{eq:dist2}
\end{equation}

\subsection{Gradient-weighted Class Activation Mapping}
As a typical gradient-based interpretability method, Gradient-weighted Class Activation Mapping (Grad-CAM) \cite{Selvaraju2016GradCAMVE} utilizes gradients to capture the salient regions of the feature maps in the final convolutional layer of DNNs, providing visual explanations for computer vision tasks.
Let \( y_c \) be the prediction score for class \( c \) (before applying the softmax activation), and \( H_k \) be the rectified feature map of a certain convolutional layer (e.g., the last one). The gradient of \( y_c \) with respect to the feature map \( H_k \) is denoted as \( \frac{\partial y_c}{\partial H_k} \).

Through backpropagation, we can compute the importance weight of each neuron using the gradient, which is given by:

\begin{equation}
\alpha_k^c = \frac{1}{M} \sum_m \sum_n \frac{\partial y_c}{\partial H_k}
\label{eq:cam}
\end{equation}

where \( M \) represents the product of the width and height of the feature map \( H_k \), \( \alpha_k^c \) denotes the importance of the feature map \( H_k \) for class \( c \), and \( H_k \) is the data of the feature map \( H \) at the coordinate \( (m, n) \) in channel \( k \).

Then, the localization map \( L_c^{\text{Grad-CAM}} \) can be obtained according to the following expression:

\begin{equation}
L_c^{\text{Grad-CAM}} = \text{ReLU} \left( \sum_k \alpha_k^c H_k \right)
\label{eq:cam_map}
\end{equation}

where the ReLU function is applied to retain the features that have a positive impact on class \( c \).

\section{Methodology}
In this section, we give the problem statement and the description of TIP.

\begin{figure}[t]
    \centering 
    \includegraphics[width=0.5\textwidth]{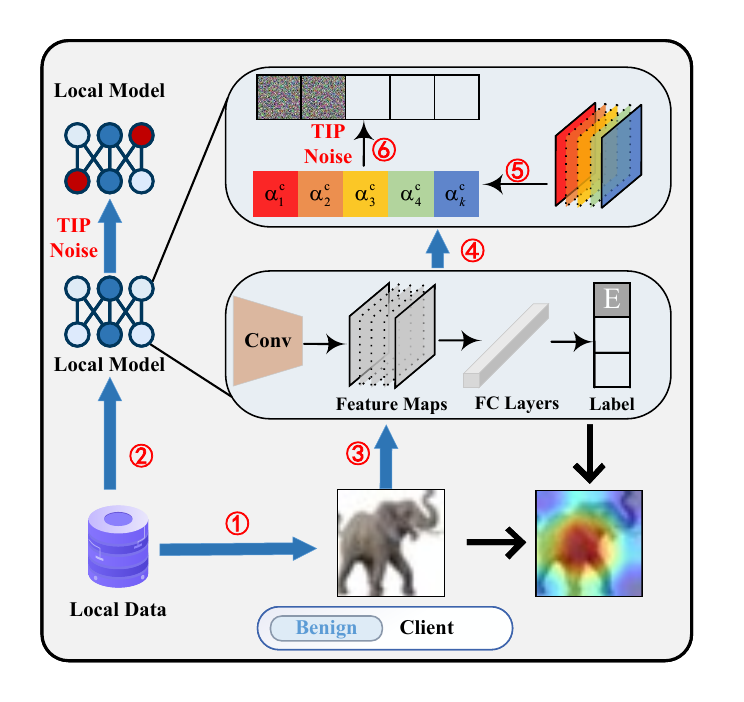}
    \caption{\centering Targeted Interpretable Perturbation in FL}
    \label{fig:TIP_FL}
\end{figure}

\subsection{Problem Formulation and Threat Model}
FL mitigates the direct sharing of raw data by enabling clients to train models locally and upload only model updates. However, prior studies have shown that gradients or model updates themselves still implicitly encode rich semantic information. Adversaries can exploit this information to conduct reconstruction-based attacks, thereby recovering sensitive training data from clients.
In this paper, we consider the following threat model: the server, or an adversary with a server-side perspective, is honest-but-curious. Such an adversary faithfully follows the FL training and aggregation protocol, but attempts to infer the original input data from the locally uploaded model parameters or gradients. The adversary cannot directly access clients’ private data, yet can observe the model updates transmitted in each communication round.

Under this threat model, a critical challenge is how to effectively suppress the re-constructable sensitive semantic information contained in model updates without significantly degrading model performance or disrupting convergence behavior.

\subsection{Method: TIP Method}
We propose the TIP mechanism, which exploits the semantic information implicitly embedded in gradient updates during FL to apply channel-level, frequency-domain selective perturbations to sensitive regions of the model, thereby defending against reconstruction-based attacks. The overall framework is illustrated in Fig. \ref{fig:TIP_FL}.

Before each communication round, the client first completes local model training without introducing noise, ensuring that the resulting gradients faithfully capture the true task semantics. Subsequently, each client selects a small subset of representative class samples from its local dataset as guidance samples. These samples are forwarded through the model to obtain logits, and the corresponding error gradients are computed via backpropagation.

On this basis, we introduce the Sensitive Importance-aware Kernel (SIK) algorithm to perform channel-wise importance analysis on convolutional layers using Grad-CAM. Specifically, Grad-CAM computes importance weights for each convolutional channel based on the error gradients. After ranking these weights, the top-k most important channels are selected as potentially sensitive channels.

For the convolutional kernel parameters corresponding to the selected channels, we transform them into the frequency domain and focus on their high-frequency components. Low-amplitude noise is injected only into the high-frequency regions, after which an inverse transformation is applied to map the perturbed parameters back to the spatial domain. This design ensures that the uploaded model updates effectively disrupt an adversary’s inference of sensitive feature pathways, while avoiding significant degradation of overall model performance.

Finally, each client uploads the locally updated model parameters with selectively perturbed high-frequency components to the server, which then performs standard aggregation to update the global model. The explanation of each step in Fig. \ref{fig:TIP_FL} is as follows:

\textbf{Step 1: Noise-Free Local Training.}
Upon receiving the global model, each client initializes its local model and performs standard local training on its private dataset without injecting any noise, ensuring that the resulting gradients faithfully encode task-related semantic information.
\begin{equation}
    \min_{\mathbf{w}} F(\mathbf{w})
    := \frac{1}{N} \sum_{n=1}^{N} 
    \ell\big(f_{\mathbf{w}}(x_n), y_n\big),
\end{equation}
where $\ell(\cdot,\cdot)$ denotes the loss function and $f_{\mathbf{w}}$ is the model parameterized by $\mathbf{w}$.

\textbf{Step 2: Guidance Sample Selection.}
Each client selects a small set of representative class samples from its local dataset as guidance samples, which are used to probe the semantic sensitivity of different model components.
Given a labeled dataset
$
\mathcal{D} = \{(x_n, y_n)\}_{n=1}^N
$
with label set $\mathcal{Y}$, we construct a representative set by randomly sampling exactly one instance from each class.
Specifically, for each class $c \in \mathcal{Y}$, let
$
I_c := \{\, n \in \{1,\dots,N\} : y_n = c \,\}.
$
We sample one index $i_c$ uniformly at random from $i_c \sim \mathrm{Unif}(I_c)$, and define the representative index set as $S := \{ i_c : c \in \mathcal{Y} \}$. The representative sample set is then given by $R := \{ x_n : n \in S \}$.By construction, the sampling strategy satisfies 
\begin{equation}
\forall c \in Y,\, |S \cap I_c| = 1
\end{equation}

\textbf{Step 3: Forward–backward Gradient Computation.} Each selected guidance sample is individually forward-propagated through the local model to obtain its logits, and then backpropagated to compute its error gradient. Where $(x,y) \in R$
\begin{equation}
\nabla_w L(w) 
= \frac{\partial \ell(f(x;w), y)}{\partial f}\,
\frac{\partial f(x;w)}{\partial w}
\end{equation}

\textbf{Step 4: Sensitive Importance-aware Kernel.}
The SIK algorithm is applied to identify sensitive convolutional channels. Specifically, Grad-CAM is used to compute channel-wise importance scores $\alpha^c$ based on error gradients, and the top-k most important channels are selected after ranking.

\textbf{Step 5: Frequency-Domain Selective Perturbation.} For the selected channels, the corresponding convolutional kernel parameters are transformed into the frequency domain using the DFT. On this basis, low-amplitude noise is injected into their high-frequency components via a masking operation, after which the perturbed parameters are mapped back to the spatial domain through an inverse transform, as shown in lines 8–17 of Algorithm \ref{alg1}.

\textbf{Step 6: Model Upload and Aggregation.} Finally, the locally trained models with selectively perturbed high-frequency parameters are uploaded to the server, where they are aggregated to produce the updated global model for the next communication round.

\subsection{Explanation of Algorithm 1: TIP}
The proposed high-frequency perturbation injection strategy is summarized in Algorithm~1. 
The set of convolutional layers selected for perturbation and their corresponding parameters are stored in $L$ and $P_t$, respectively. 
Line~2 computes the importance scores of the channels in each target convolutional layer using Grad-CAM with class-representative samples $R$.

In Line 3, the top $N_r$ channels are selected according to the proportion \( r \), based on the sorted importance scores in descending order. Here, the proportion \( r \) is set to 0.1.
In Line 4, Select the top $N_r$ channels based on the importance scores for noise injection.

Lines~5--6 construct a binary high-frequency preservation mask $I_{\text{high}}^{(R)}$ in the frequency domain according to the distance from the spectral center, retaining only components outside the radius $R$. In this paper, the radius $R$ is set to 0.5.
Lines 8–13 follow the differential privacy framework to determine the perturbation strength. The noise scale \( \zeta \) is calculated as:
$\zeta = \frac{b \Delta s}{\epsilon}$
where \( b \geq \sqrt{2 \ln \left( \frac{1.25}{\delta} \right)} \) for \( \delta \in (0, 1) \), and \( \Delta s \) denotes the sensitivity. The resulting noise scale serves as a global noise control parameter.In implementation, we adopt a layer-wise noise injection strategy: standard Gaussian noise $\mathcal{N}(0,1)$, s first sampled and transformed into the frequency domain, where a high-frequency mask is applied. The DP-calibrated global standard deviation is then used as a noise control parameter and evenly distributed across network layers. Specifically, the noise at each layer is scaled by $\beta = \frac{\zeta}{\left|L_c\right|}$, where $\left|L_c\right|$ denotes the number of perturbed layers. We set $0 < \beta \leq 1$.

In Line~15, the convolutional weights of each channel are transformed into the frequency domain via a two-dimensional DFT. 
Line~16 adds the masked high-frequency noise, to the frequency-domain weights. 
Line~17 applies the inverse DFT to recover the perturbed parameters in the spatial domain. 
Finally, in Line~21, the perturbed local parameters $\tilde{w}_i$ are uploaded to the central server for aggregation.

\begin{algorithm}[t]
\caption{Targeted Interpretable Perturbation.}\label{alg:alg1}
\textbf{Input:} 
$R$: High-frequency preservation mask radius, $L_c$: Convolutional layers with noise injection, $r$: Proportion of important channels, $N$: Number of channels, $C_s$: Classes representative sample. $\beta$ denotes the high-frequency noise scaling factor.

\textbf{Output:} 
 $\tilde{w}_i$: The perturbed local model parameters are uploaded to the central server for aggregation.
\begin{algorithmic}[1]
\FOR{each layer $l \in L_c$}
    \STATE Calculate the importance of channel weights: $\alpha^c = \text{GradCAM}(w_i, l, C_s)$
    \STATE Compute the number of important channels: $N_r = N \cdot r$
    \STATE \textbf{SIK:} Select the indices of the top $N_r$ channels based on their importance scores: $\alpha^{\text{top}} = \text{TopN}(\alpha^c, N_r)$
    \FOR{each kernel $k^{C\times H\times W}$ in $l$}
        \STATE Build a high-frequency binary mask: $\displaystyle 
        I_{\text{high}}^{(R)}(u,v) =
        \begin{cases}
        1, & (u-H/2)^2 + (v-W/2)^2 \ge R^2 \\
        0, & (u-H/2)^2 + (v-W/2)^2 < R^2
        \end{cases}$ 
        \FOR{each channel $c^{H\times W}$ in $k^{C\times H\times W}$}
            \IF{$k \in \alpha^{\text{top}}$}
                \STATE Generate Gaussian noise: ${\text{noise}}_g^{H\times W} \sim \mathcal{N}(0, 1)$
                \STATE Transform noise to frequency domain:
                $F_g = \text{FFT}({\text{noise}}_g^{H\times W})$
                \STATE Apply high-frequency mask: ${\text{Noise}}_{l,k,c} = F_g \cdot I_{\text{high}}^{(R)} \cdot \beta$
            \ELSE
                \STATE ${\text{Noise}}_{l,k,c} = \text{Zeros}(H,W)$
            \ENDIF
            \STATE Transform channel weights to frequency domain: $F_c = \text{FFT}(c^{H\times W})$
            \STATE Inject perturbation: $F_c^n = F_c +{\text{Noise}}_{l,k,c}$
            \STATE Convert back to spatial domain: $c^{H\times W} = \text{Re}(F_c^n)$
        \ENDFOR
    \ENDFOR
\ENDFOR
\STATE Submit perturbed local parameters: $\tilde{w}_i$
\end{algorithmic}
\label{alg1}
\end{algorithm}

\begin{table*}[htbp]
\centering
\caption{Quantitative Evaluation for Different Attack Methods}
\begin{tabular}{llcccccccc}
\toprule
\multirow{2}{*}{} & \multirow{2}{*}{\textbf{Datasets}} &
\multicolumn{4}{c}{\textbf{IG}} & \multicolumn{4}{c}{\textbf{GIFD}} \\
\cmidrule(r){3-6}\cmidrule(l){7-10}
& & \textbf{No DP} & \textbf{DP} & \textbf{AGP} & \textbf{TIP} & \textbf{No DP} & \textbf{DP} & \textbf{AGP} & \textbf{TIP} \\
\midrule
\multirow{3}{*}{\textbf{PNSR↓}}
& CIFAR-10       & 14.58 & 9.18 & 9.54 & 9.85 & 20.4  & 10.33 & 10.3 & 11.15 \\
& CIFAR-100      & 14.10 & 8.20 & 7.93 & 8.73 & 21.39 & 9.40  & 10.3 & 10.70 \\
& TINY-IMAGENET & 12.20 & 6.73 & 6.77 & 6.96 & 14.31 & 8.67  & 8.63 & 9.09  \\
\midrule
\multirow{3}{*}{\textbf{SSIM↓}}
& CIFAR-10       & 0.37  & 0.02  & 0.03  & 0.01  & 0.57 & 0.05 & 0.04 & 0.04 \\
& CIFAR-100      & 0.30  & 0.005 & 0.001 & 0.01  & 0.47 & 0.02 & 0.04 & 0.04 \\
& TINY-IMAGENET & 0.17  & 0.002 & 0.005 & 0.009 & 0.24 & 0.02 & 0.02 & 0.02 \\
\bottomrule
\label{tav:recon}
\end{tabular}
\end{table*}

\begin{table*}[htbp]
\centering
\caption{Quantitative Comparison on Different Datasets}
\label{tab:quantitative_results}
\begin{tabular}{lccc|ccc|ccc}
\toprule
\textbf{Datasets} & \multicolumn{3}{c|}{\textbf{MSE↓}} & \multicolumn{3}{c|}{\textbf{PSNR↑}} & \multicolumn{3}{c}{\textbf{SSIM↑}} \\
\midrule
& \textbf{DP} & \textbf{APG} & \textbf{TIP} & \textbf{DP} & \textbf{APG} & \textbf{TIP} & \textbf{DP} & \textbf{APG} & \textbf{TIP} \\
\midrule
CIFAR-10 & 5044.96 & 3215.65 & 16.97 & 11.18 & 13.47 & 39.50 & 0.35 & 0.48 & 0.99 \\
CIFAR-100 & 4948.12 & 3045.79 & 46.35 & 11.28 & 13.82 & 32.48 & 0.32 & 0.51 & 0.98 \\
TINY-IMAGENET & 5104.48 & 4039.55 & 753.50 & 11.27 & 12.39 & 22.03 & 0.60 & 0.66 & 0.91 \\
\bottomrule
\end{tabular}
\end{table*}

\section{Results and Analysis}

In this section, we present the experimental setup and then report the evaluation results. We conduct extensive experiments to assess the performance of the proposed TIP in fl. First, we examine the impact of TIP on gradient-based interpretability. Then, we investigate its effect on classification accuracy. Next, we verify the robustness of TIP against privacy leakage attacks. Finally, we perform convergence analysis and discuss the computational cost.

\subsection{Model Configuration and Parameter Settings}
We adopt ResNet18 as the image classification model, which consists of 17 convolutional layers and one fully connected layer, and use SGD as the optimizer with a learning rate of 0.001. The weights of all convolutional layers are selected as noise-injected targets. To evaluate the practical convergence performance of TIP under the FedAvg framework, the total number of clients is set to 100, with 10\% of clients randomly selected to participate in each communication round. The number of communication rounds is set to 1000. The local training batch size is set to 32, and each client trains for 2 epochs per round, with the privacy budget $\epsilon$ fixed to 5 for each client in every communication round. The radius of the high-frequency mask $R$ is set to 0.5, and the proportion of important channels is set to 10\%. For interpretability comparison experiments involving IG and GIFD, the batch size is additionally set to 1, and the number of epochs is set to 4.

\subsection{Datasets}
\begin{itemize}
\item{CIFAR-10. CIFAR-10 \cite{Krizhevsky2009LearningML} is made up of 10 classes of 32x32 images with three RGB channels and consists of 50,000 training samples and 10,000 testing samples.}
\item{CIFAR-100. Similarly, CIFAR-100 \cite{Krizhevsky2009LearningML} has the same image size. However, it exists 100 classes with 600 samples (500 training samples and 100 test samples) in each class. In other words, CIFAR-100 is a more sophisticated dataset than CIFAR-10 to evaluate the FL system fairly}.
\item{TINY-IMAGENET-200. TINY-IMAGENET-200 is a subset of the ImageNet\cite{deng2009imagenet} dataset, consisting of 200 object classes. Each class contains 500 training images, 50 validation images, and 50 test images. All images are resized to a resolution of 64×64 pixels.}
\end{itemize}

\subsection{Evaluation Metrics}
We evaluate the performance of our method both qualitatively and quantitatively. First, we use Grad-CAM to generate interpretable visualizations for qualitative analysis of interpretability. Then, we used Mean Squared Error (MSE), Structural Similarity Index (SSIM), and Peak Signal-to-Noise Ratio (PSNR) to quantitatively assess the similarity between visualized heatmaps without using DP and those using DP, AGP, and TIP, as well as the similarity between the original images and the recovered images under GIAs.
The detailed calculation is as follows, where $X$ and $Y$ denote the two images being compared. For MSE
\begin{equation}
    \mathrm{MSE}(X,Y)=\frac{1}{3HW}\sum_{c=1}^{3}\sum_{i=1}^{H}\sum_{j=1}^{W}(X_{ijc}-Y_{ijc})^{2}
\end{equation}
Than for PNSR 
\begin{equation}
\mathrm{PSNR}(X,Y)=10\log_{10}\left(\frac{\mathrm{MAX}^2}{\mathrm{MSE}(X,Y)}\right)
\end{equation}
Finally, for the calculation of SSIM.
\begin{equation}
\begin{aligned}
\\
\mathrm{SSIM}(X,Y)=\frac{1}{3}\sum_{c\in\{R,G,B\}}\mathrm{SSIM}\bigl(X^{(c)},Y^{(c)}\bigr)
\\
\mathrm{SSIM}(U,V)
=
\frac{(2\mu_U\mu_V + C_1)(2\sigma_{UV} + C_2)}
{(\mu_U^2 + \mu_V^2 + C_1)(\sigma_U^2 + \sigma_V^2 + C_2)}
\\
C_1 = (K_1 \mathrm{MAX})^2, \quad
C_2 = (K_2 \mathrm{MAX})^2
\\
\text{Typical values: } K_1 = 0.01,\quad K_2 = 0.03
\end{aligned}
\end{equation}

\subsection{Baselines}
To comprehensively evaluate the effectiveness of the proposed defense, we consider both defense-oriented and attack-oriented baselines.

\subsubsection{Defense baselines.}
\begin{itemize}
    \item No DP. This setting corresponds to the standard FedAvg framework without any DP mechanism applied. It serves as a fundamental reference to assess the intrinsic vulnerability of the model and to provide a comparison point for privacy-preserving defenses.
    \item DP. Differential privacy is applied on top of the No DP setting by adding Gaussian noise to the model parameters during training. This represents a commonly used DP-based defense baseline in FL.
    \item APG. APG\cite{Li2024TowardsAP} is an additional defense baseline that defends against information leakage in gradients by adaptively injecting DP noise into selected parameters. Compared to pure DP noise injection, APG selectively adds DP noise to the target layers, while the non-target layers remain consistent with pure DP. It serves as a competitive baseline against the defense method that we propose.
\end{itemize}

\subsubsection{Attack baselines.}
\begin{itemize}
    \item \textbf{IG}. The IG attack proposed by Geiping et al. \cite{hu2024learn} is adopted as a strong GIA baseline. IG reconstructs private training data by iteratively optimizing dummy inputs such that the gradients generated by these inputs closely match the shared gradients uploaded by clients. In addition, IG incorporates image priors to enhance the visual quality of the reconstructed samples, enabling effective data recovery even in DNNs.
    \item GIFD. GIFD \cite{fang2023gifd} introduces generative adversarial networks (GANs) as prior guidance and innovatively decomposes the GAN model to optimize the search in the intermediate feature space, thereby significantly enhancing the effectiveness of existing GIAs.
\end{itemize}

These baselines enable a systematic evaluation of the proposed defense under different privacy-preserving mechanisms and gradient-based interpretability attacks.

\subsection{Impact of AGP on Interpretability}
First, we explore the impact of our AGP on the interpretability of classification results using three benchmark datasets, as shown in Fig.4 and Fig.5. 

In this experiment, we set the ratio of channels identified as important for TIP and AGP to 0.1 and 0.25, respectively, meaning that 10\% noise is injected into the target parameters of TIP, and 25\% noise is injected into the target parameters of AGP, while 100\% noise is injected into the non-target parameters of AGP. To clearly demonstrate the effect of these methods, we present the visualizations of AGP and TIP heatmaps. 

Regarding the privacy budget, we balance accuracy and privacy protection by setting it to 5 and 1 for AGP and TIP methods. Fig.4 shows the visual explanations of different methods using heatmaps under the IID scenario. We first focus on the visualizations for the CIFAR-10 and CIFAR-100 datasets. It can be observed that DP no longer accurately localizes the key parts of the images, leading to poor interpretability. In contrast, our TIP still accurately localizes the most recognizable parts of the center of each image, thereby enhancing interpretability and recognition ability. For example, in the first four images in Fig. 4, TIP is able to identify most of the contours that significantly contribute to the prediction, compared to AGP. 

Additionally, the red area is more comprehensive, prominent, and similar to that of the No DP case. This indicates that TIP provides better interpretability of the classification results. We further compare and explain the visualizations on the Tiny ImageNet-200 dataset. As shown in the last two rows of images in Fig. 5, it can be observed from the heatmaps that our TIP better maintains the focus area consistent with the naïve training model compared to AGP and even without DP. This suggests that our method causes minimal disruption to the model's interpretability. Therefore, it has the potential to accelerate the deployment of FL in safety-critical applications. 
To further quantitatively evaluate interpretability, we use MSE , SSIM , and PSNR  to compute the similarity between the visualized heatmaps without using DP and those using DP, AGP, and TIP. Table. \ref{tab:quantitative_results}. shows the quantitative results of interpretable visualizations in Fig. 4. Here, we select 10 representative images from each dataset to assess their interpretability under IID. We observe that our TIP improves the interpretability performance in terms of SSIM and PSNR.

\subsection{Robustness Against GIAs}
We further demonstrate the robustness of the proposed TIP against GIAs. We adopt the well-known algorithm IG  and its improved variant GIFD  to attack our method. Our goal is to verify whether these attack methods can reconstruct the input data based on the shared weights from local clients. 

To rigorously test the robustness of our method against gradient leakage attacks, we have created an environment conducive to enhancing attack effectiveness, even though some settings may not be realistic and will be more complex in actual systems \cite{Wang2023MoreTE}. Specifically, during the model training phase of FL, we did not use momentum, weight decay, or learning rate schedulers. During the IG and GIFD phases, we set the batch size for the recovery of images to 1 and limited the training to 1 epoch. 

Fig. 5. illustrates the reconstruction of input data by the attackers under different methods. We can see that both IG and GIFD can successfully recover the input data from the shared weights, but they fail under TIP, where the recovered images are highly noisy or the image details are severely distorted, making them unrecognizable. The primary reason is that we add sufficient Gaussian noise to model parameters in our TIP, as mentioned in \cite{Zhu2019DeepLF}. When the noise is large enough, the effectiveness of the attackers is significantly reduced. 

Therefore, from a visualization perspective, TIP achieves a similar level of privacy protection as DP. Additionally, we use SSIM (lower is better) and PSNR (lower is better) to calculate the similarity between the original images and the recovered images under GIAs. Table II shows the average SSIM and PSNR values for 10 representative images from each dataset. The SSIM values for images recovered using TIP are nearly identical to those using DP, and the PSNR values are comparable, occasionally even outperforming DP.

Compared to the substantial gap between No DP and both DP and TIP, the difference between TIP and DP is negligible. This further confirms that TIP can achieve comparable privacy protection to DP, providing sufficient protection. Based on these experimental results, we can conclude that the images recovered by attackers under TIP differ significantly from the original ones both visually and quantitatively, demonstrating that our TIP is robust against IG and GIFD, thus sufficiently protecting private data in FL.

\begin{figure}[h]
  \centering
  \includegraphics[width=\linewidth]{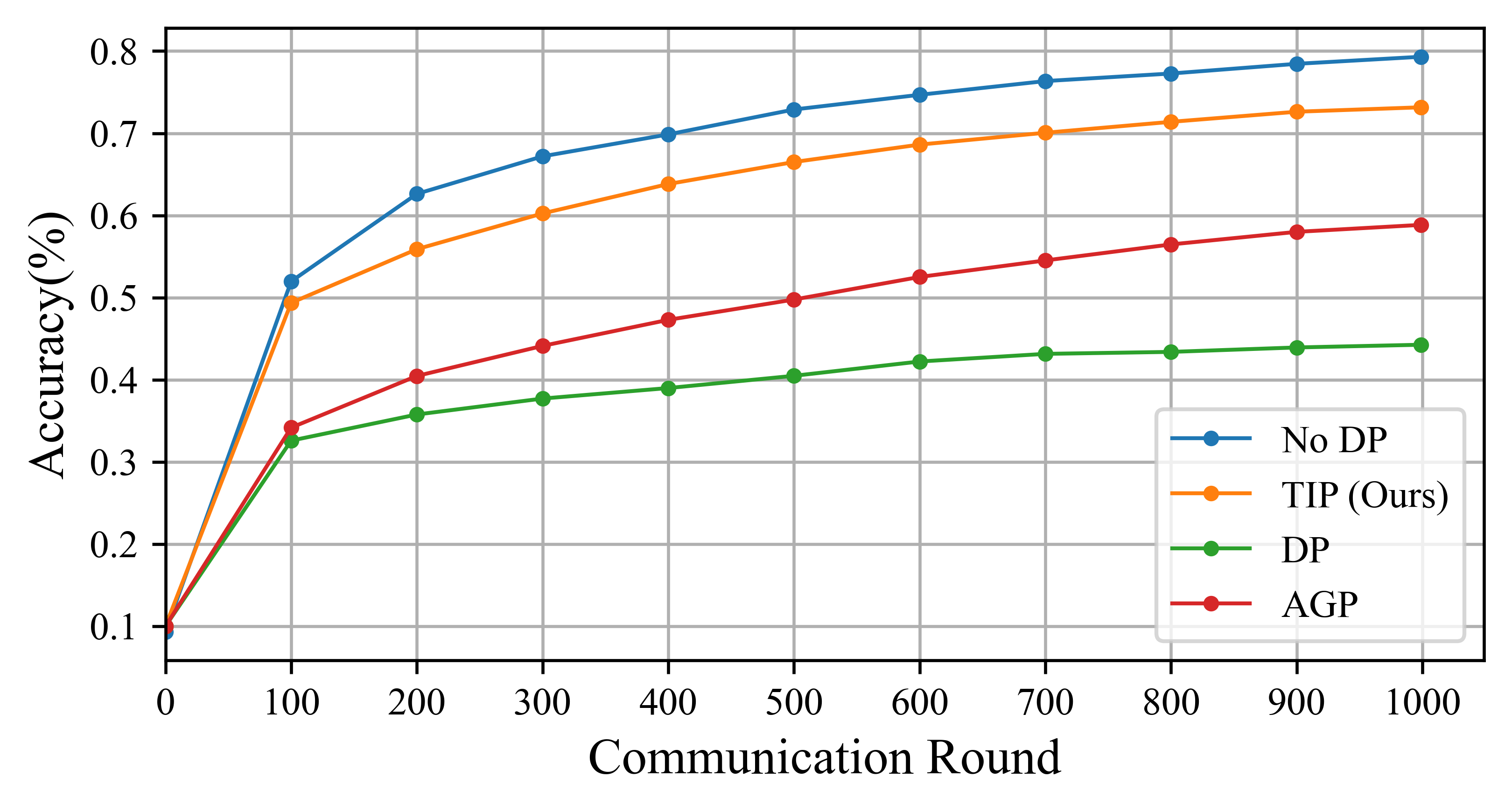}
  \caption{The Convergence ACC Plot.}
  \Description{}
\label{fig:Convergence}
\end{figure}

\subsection{Convergence}
We present the convergence performance of the proposed TIP on the CIFAR-10 dataset under the IID scenario, comparing it with No DP and DP, as shown in Figure \ref{fig:Convergence}. As observed, TIP achieves higher accuracy than DP within the same number of communication rounds in both scenarios. Furthermore, to reach the same level of accuracy, TIP requires fewer communication rounds. These results indicate that TIP converges well, consistent with our theoretical analysis. The convergence proof of the theory is shown in the Appendix.A.

\section{Conclusion}

In this paper, we proposed Targeted Interpretable Perturbation (TIP), a defense framework designed to reconcile the conflict between privacy and utility in Federated Learning. By integrating Gradient-weighted Class Activation Mapping (Grad-CAM) sensitivity analysis with frequency-domain perturbation, TIP selectively disrupts the high-frequency details essential for Gradient Inversion Attack (GIA) while preserving the low-frequency semantics critical for model generalization. Extensive experiments confirm that TIP effectively thwarts advanced GIAs while maintaining model accuracy comparable to non-private baselines. Supported by theoretical convergence guarantees, TIP offers a robust solution for secure FL and paves the way for future interpretability-guided defenses against broader adversarial threats.




\bibliographystyle{unsrt} 
\bibliography{ref} 


\appendix

\section{Convergence Guarantee}
Inspired by \cite{Li2019OnTC}, \cite{Li2024TowardsAP}, we theoretically analyze the convergence of the widely used FedAvg algorithm with our proposed TIP based on the following five \textbf{Assumptions}.

\textbf{Assumption} 1: $F_1,\dots,F_j$ are all L-smooth: for all $v$ and $w$, $\| \nabla F_j(v) - \nabla F_j(w) \| \leq L \| v - w \|$, $F_j(v) \leq F_j(w) + (v - w)^T \nabla F_j(w) + \frac{L}{2} \| v - w \|^2$.

\textbf{Assumption} 2: $F_1,\dots,F_j$ are all $\mu$-strongly convex: for all $v$ and $w$, $\| \nabla F_j(v) - \nabla F_j(w) \| \leq L \| v - w \|$, $F_j(v) \geq F_j(w) + (v - w)^T \nabla F_j(w) + \frac{\mu}{2} \| v - w \|^2$

\textbf{Assumption} 3: Let $\xi_j^t$ be sampled from the $j$-th device’s local data uniformly at random. The variance of 
stochastic gradients in each device is bounded: $\mathbb{E} \left\| \nabla F_j(w_{j}^t, \xi_{j}^t) - \nabla F_j(w_{j}^t) \right\|^2 \leq \sigma_j^2 \quad \text{for} \quad j = 1, \ldots, J$.

\textbf{Assumption} 4: The expected squared norm of stochastic gradients is uniformly bounded, i.e.,$\mathbb{E} \left\| \nabla F_j(w_{j}^t, \xi_{j}^t) \right\|^2 \leq G^2 \quad \text{for all} \quad j = 1, \ldots, J \quad \text{and} \quad t = 1, \ldots, T - 1$.

\textbf{Assumption} 5: Assume $S_t$ contains a subset of $N$ indices
 uniformly sampled from $[J]$ without replacement. Assume the
data are balanced in the sense that $p_1=\dots = p_J = \frac{1}{J}$.The aggregation step of FedAvg performs $w_t \leftarrow \frac{J}{N} \sum_{j \in S_t} p_j w_j^t$.

We define $F^\ast$ and $F^\ast_j$ as the minimum value of $F$ and $F_j$ respectively \cite{Li2019OnTC}. As mentioned in Section \ref{sec:preli_fed} $F_1,\dots F_J$ represent local objective function for each client.
The quantity $\Gamma = F^\ast - \sum_{j=1}^{J} p_j F_j^\ast$ serves as a measure of data inconsistency across clients. A larger value of $\Gamma$ indicates stronger distributional divergence among local datasets. Under an IID setting, the discrepancy vanishes as the number of samples increases, and thus $\Gamma$ approaches zero.

\textbf{Theorem} 1: Suppose that \textbf{Assumptions} 1–5 are satisfied, and let $L,\mu ,{\sigma _j},G$ be defined accordingly. Define $\theta  = \frac{\mu }{L},\phi  = \max \{ 8\theta ,E\} $, and adopt the step size schedule ${\eta _t} = \frac{2}{{\mu (\phi  + t)}}$. Therefore, it can be concluded that

\begin{equation}
\begin{aligned}
&\mathbb{E}\big[F(w_T)\big] - F^\ast
\\
&\le\; \frac{\theta}{\gamma + T}
\left(
    \frac{2(B+C)}{\mu}
    + \frac{\mu\phi}{2}\mathbb{E}\|w_0 - w^\ast\|^2
\right),
\end{aligned}
\label{eq:convergence}
\end{equation}

 the constants $B$ and $C$ are given by
\begin{equation}
\label{eq:b}
\begin{aligned}
B = \sum\limits_{j = 1}^J {p_j^2} \left[ {(r{K}){L^2}\zeta _{{\rm{eff}}}^2 + \sigma _j^2} \right] + 6LT
\\
+ 8{(E - 1)^2}\left[ {(r{K}){L^2}\zeta _{{\rm{eff}}}^2 + {G^2}} \right]
\\
C = \frac{{J - N}}{{J - 1}} \cdot \frac{4}{N}{E^2}\left[ {(r{K}){L^2}\zeta _{{\rm{eff}}}^2 + {G^2}} \right]
\end{aligned}
\end{equation}
Where $ K$ is the total number of channels (parameters) in the considered model layer for each client, is the proportion of channels deemed important and thus injected with frequency-domain high-frequency noise, $\zeta _{{\rm{eff}}}^2$ denotes the effective variance of the noise in the time domain after inverse frequency transform retaining only high-frequency components.

\begin{proof}
For a specific channel $\kappa$ within the subset of important channels selected for perturbation, the noise injection operation can be expressed as:

\begin{equation}
\label{eq:proof_weight_noise}
\begin{aligned}
{\tilde w_{j,\kappa }^t} = {\mathcal{F}^{ - 1}}\left( {\mathcal{F}({w_{j,\kappa }^t}) + I_{high}^{R} \odot {{\hat n}_\kappa }} \cdot \beta\right)
\end{aligned}
\end{equation}

where ${\hat n_\kappa } = {\mathcal F}({n_\kappa })$ and $n_\kappa \sim \mathcal{N}\!\left(0,\;\zeta^2 I\right)$. Here, ${\mathcal F}$ and ${{\mathcal F}^{ - 1}}$ denote the DFT and its inverse, respectively, $I_{high}^{R}$ is a binary mask selecting the high-frequency components, and represents element-wise multiplication. By Parseval's theorem (assuming unitary DFT normalization), the time-domain perturbation
\begin{equation}
\label{ time-domain perturbation}
\begin{aligned}
{\delta _\kappa } = {\tilde w_{j,\kappa }^t} - {w_{j,\kappa }^t}
\end{aligned}
\end{equation}
Satisfies
\begin{equation}
\label{time_domain_perturbation_Satisfies}
\begin{aligned}
\mathbb{E}\!\left\lVert \delta_k \right\rVert^2
= \mathbb{E}\!\left\lVert \beta \mathbf{H} \odot \hat{n}_k \right\rVert^2
= \beta^2\zeta_{\mathrm{eff}}^{\,2} \leq \zeta_{\mathrm{eff}}^{\,2}
\end{aligned}
\end{equation}
Since only the $r{K_t}$ important channels are noised, each with $\zeta _{{\rm{eff}}}^2$:

\begin{equation}
\begin{aligned}
\mathbb{E}\!\left\lVert \tilde{W}_{j}^t - W_{j}^t \right\rVert^2
= r K \zeta_{\mathrm{eff}}^{\,2}
\end{aligned}
\end{equation}
Thus, we obtain:

\begin{equation}
\begin{aligned}
\label{eq:res}
&\mathbb{E}\!\left\lVert\nabla F_j\!\left(\tilde{W}_{j}^t, \xi_{j}^t\right)-\nabla F_j\!\left(W_{j}^t, \xi_{j}^t\right)\right\rVert^2 
\\
&= \sum_{k \in K}\mathbb{E}\!\left\lVert\nabla F_j\!\left(\tilde{w}_{jk}^{\,t}, \xi_j^{\,t}\right)-\nabla F_j\!\left(w_{jk}^{\,t}, \xi_j^{\,t}\right) \right\rVert^2
\\
&\leq{(r{K}){L^2}\zeta _{{\rm{eff}}}^2}
\end{aligned}
\end{equation}

According to the norm triangle inequality, we leverage \textbf{Assumption} 3 and \ref{eq:res} to obtain a new bound on the variance of stochastic gradients in each client as follows:

\begin{equation}
\begin{aligned}
&\mathbb{E}\!\left\lVert
\nabla F_j\!\left(\tilde{W}_j^{\,t}, \xi_j^{\,t}\right)
-\nabla F_j\!\left(W_j^{\,t}\right)
\right\rVert^2
\\
&\leq
\mathbb{E}\!\left\lVert
\nabla F_j\!\left(\tilde{W}_j^{\,t}, \xi_j^{\,t}\right)
-\nabla F_j\!\left(W_j^{\,t}, \xi_j^{\,t}\right)
\right\rVert^2
\\
&\quad+
\mathbb{E}\!\left\lVert
\nabla F_j\!\left(W_j^{\,t}, \xi_j^{\,t}\right)
-\nabla F_j\!\left(W_j^{\,t}\right)
\right\rVert^2
\\
&\leq{(r{K}){L^2}\zeta _{{\rm{eff}}}^2} + \sigma_j^2
\end{aligned}
\end{equation}
Similarly, the new bound for \textbf{Assumption} 4 with our  method can be derived as:$\mathbb{E} \left\| \nabla F_j(w_{j}^t, \xi_{j}^t) \right\|^2 \leq (r{K}){L^2}\zeta _{{\rm{eff}}}^2 + G^2$

By applying the new bounds for \textbf{Assumptions} 3 and 4, the convergence guarantee with our proposed TIP can be obtained as equation \ref{eq:convergence}. \end{proof}

\end{document}